\documentclass[runningheads]{llncs}
\usepackage{graphicx}
\usepackage{amsmath}
\usepackage{makecell}
\usepackage{float}
\begin{document}
\title{Predicting Deterioration in Mild Cognitive Impairment with Survival Transformers, Extreme Gradient Boosting and Cox Proportional Hazard Modelling}
\titlerunning{Predicting Deterioration in MCI with Survival Transformers.}
%
\author{Henry Musto\inst{1} \and
Daniel Stamate\inst{1,2} \and
Doina Logofatu\inst{3} \and
Daniel Stahl\inst{4}}
\authorrunning{H. Musto et al.}
%
\institute{Data Science and Soft Compuputing Lab, Goldsmiths, University of London, London, United Kingdom
\email{hthom018@gold.ac.uk}\\ \and
School of Health Sciences, The University of Manchester, Manchester, United Kingdom\\ \and
Faculty of Computer Science and Engineering, Frankfurt University of Applied Sciences, Frankfurt, Germany\\ \and
Department of Biostatistics, Institute of Psychiatry, Psychology and Neuroscience King’s College London, London, United Kingdom
}

\maketitle

\begin{abstract}
The paper proposes a novel approach of survival transformers and extreme gradient boosting models in predicting cognitive deterioration in individuals with mild cognitive impairment (MCI) using metabolomics data in the ADNI cohort. By leveraging advanced machine learning and transformer-based techniques applied in survival analysis, the proposed approach highlights the potential of these techniques for more accurate early detection and intervention in Alzheimer’s dementia disease.
This research also underscores the importance of non-invasive biomarkers and innovative modelling tools in enhancing the accuracy of dementia risk assessments, offering new avenues for clinical practice and patient care. A comprehensive Monte Carlo simulation procedure consisting of 100 repetitions of a nested cross-validation in which models were trained and evaluated, indicates that the survival machine learning models based on Transformer and XGBoost achieved the highest mean C-index performances, namely 0.85 and 0.8, respectively, and that they are superior to the conventional survival analysis Cox Proportional Hazards model which achieved a mean C-Index of  0.77. Moreover, based on the standard deviations of the C-Index performances obtained in the Monte Carlo simulation, we established that both survival machine learning models above are more stable than the conventional statistical model.
\end{abstract}
\section{Introduction}
Dementia diseases represent one of the most significant global health challenges of our time, affecting millions worldwide with profound impacts on individuals, families, and healthcare systems \cite{international_world_2023}. Alzheimer's Disease (AD) is the most common form of dementia, with between 60\%-80\% of dementia being of AD type \cite{noauthor_dementia_nodate}. Early detection and intervention are crucial for slowing progression, enhancing the quality of life, and planning care \cite{stamate_predicting_2023}. However, accurately predicting cognitive deterioration in AD remains a formidable challenge, limited by the sensitivity and specificity of traditional biomarkers and cognitive assessments \cite{stamate_metabolitebased_2019}. Furthermore, the collection of these biomarkers is often costly and invasive for patients. Recent advances in blood metabolomics have opened new avenues for understanding the complex biological underpinnings of dementia. This approach offers a comprehensive snapshot of metabolic alterations, providing insights into disease onset and progression, whilst also reducing the need for invasive procedures such as lumbar punctures and neuroimaging \cite{guo_plasma_2024}. Nevertheless, the complexity and high-dimensionality of metabolomics data necessitate sophisticated analytical tools to identify potential metabolomic candidates for the creation of models for AD prediction \cite{qiang_plasma_2024} \cite{machado-fragua_circulating_2022}. Machine learning and deep learning (hereafter collectively referred to as ML) have been proposed as such analytical tools.

However, one of the challenges that prevent the successful integration of ML models into clinical practice concerns the information that can be gleaned from these methods. The techniques often deliver a binary or multinomial prediction indicating the likelihood of the development of disease. However, in prognostic modelling, it is usually more valuable to model the risk of developing the disease as a function of time. This limitation of traditional ML techniques restricts clinicians’ ability to accurately track and communicate the risk of disease occurrence over time with the patient \cite{musto_predicting_2023}. As a result, an emerging field of exploration seeks to build on classical time-dependent models, such as survival analysis, to develop machine learning models which can predict the time-dependent risk of developing AD and thus move beyond simple classification \cite{musto_survival_2023}\cite{spooner_comparison_2020}\cite{stamate_predicting_2023}\cite{mirabnahrazam_predicting_2023}.

Recently, a number of survival-based neural network models have been proposed. These models can estimate the patient-specific survival probabilities over time, by using either fully connected or recurrent neural networks \cite{wiegrebe_deep_2024}. However, recent work has found that they do not perform as well as tree-based survival models \cite{musto_predicting_2023}\cite{musto_survival_2023}. Neural networks require a high degree of computational power in order to train what are often complex models and thus the models must be shown to be superior to simpler alternatives in order to justify the additional computational cost. In the last few years the transformer model \cite{vaswani_attention_2023} has risen to prominence in the domain of Natural Language Processing as a tool for the handling sequential data, with the model using an attention mechanism that can observe more important aspects of the sequence, which are retained to help the model make sense of the whole sequence. Although, the most common application of these transformer models has been in algorithms that underpin chatbots such as ChatGPT,  because of the high degree of accuracy handling sequential data, a small number of papers have now developed transformer models for use in survival prediction in disease \cite{hu_transformer-based_nodate}\cite{wang_survtrace_2022}\cite{ding_pathology-and-genomics_2023}. 

To that end, this paper introduces a novel application of one survival transformer model to predict the risk of cognitive deterioration in AD using blood metabolomics data. By leveraging this approach, we aim to uncover the subtle metabolic shifts that precede cognitive decline, offering a more accurate and timely prediction of risk in AD. To our knowledge, this paper is the first attempt to assess the performance of survival transformers in predicting deterioration in AD using metabolomics data.

\section{Related Work}

Transformer models, originally developed for natural language processing \cite{vaswani_attention_2023}, have been adapted for survival analysis due to their ability to handle sequential data and capture long-range dependencies. Studies such as \cite{hu_transformer-based_nodate}\cite{wang_survtrace_2022} have demonstrated evidence of transformer-based models having superior predictive ability when comparing with standard Cox Proportional Hazards (Cox PH) models, and other survival-based machine learning and deep learning techniques. Furthermore, \cite{ding_pathology-and-genomics_2023} demonstrated good predictive ability when integrating high-dimensional, multimodal medical data, however they did not directly compare the model to a standard Cox PH model or any survival-based ML model.

The aforementioned studies suggest that specific survival-based transformer models may produce good predictions of temporal risk. However, the literature remains sparse and more work is needed to directly compare survival transformers and existing survival models. Furthermore, the complexity of transformer architectures introduces challenges in interpretability and computational efficiency. The opaque nature of transformer processes makes it difficult to decipher the model's decision-making process, a significant drawback in clinical settings where understanding the rationale behind predictions is crucial.

In contrast, XGBoost models have been highlighted for their robust performance in survival analysis \cite{musto_survival_2023}, particularly when dealing with structured, tabular datasets. Other works by \cite{billichova_comparing_2024} and \cite{spooner_comparison_2020} underscore survival tree-based method's competitive accuracy and computational efficiency, making them a viable option for rapid model training and deployment. Furthermore, recent work has also demonstrated the superiority of tree-based methods when compared to neural network based techniques in survival prediction tasks \cite{musto_survival_2023}\cite{musto_predicting_2023}. This is fortunate as tree-based methods are usually not as complex or computationally intensive as neural network models, making them potentially a better option when considering adoption into healthcare infrastructure and processes. One the other hand, XGBoost can also suffer from a lack of interpretibility and thus efforts must be made to allow clinicians the opportunity for inference as well as prediction when using these models. Recent efforts to implement explainability into survival ML models have demonstrated some success in allowing inference from these 'black box' models \cite{sarica_explainability_2023}. 

Recent work has demonstrated evidence for the efficacy of survival-based ML models. However, to date, no direct comparison of these models with survival transformers has been made within the context of AD prediction. 

\section{Methodology}
\subsection{Alzheimer's Disease Neuroimaging Initiative}
The data used in this paper was derived from the Alzheimer’s Disease Neuroimaging Initiative (ADNI) database \cite{noauthor_adni_nodate}. This longitudinal case-control study was initiated in 2004 by the National Institute of Aging (NIA), The National Institute of Biomedical Imaging and Bioengineering (NIBIB), The Food and Drug Administration (FDA), as well as elements of the private and non-profit sectors. The initial protocol, ADNI1, was conducted over six years, recruiting 400 participants diagnosed with Mild Cognitive Impairment (MCI), 200 participants with Alzheimer’s (AD), and 200 healthy controls (CN). The initial goal of the ADNI study was to test whether repeated collections of neuroimaging, biomarker, genetic, and clinical and neuropsychological data could be combined to contribute in an impactful way to research dementia \cite{noauthor_adni_nodate}. Data for the current study also included lipidomics sample data collected from ADNI1 participants through blood tests in a manner described in \cite{huynh_lipidomic_nodate}. Data for the present paper was downloaded on the 6th January 2024 through the ADNIMERGE package in R. This package combines predictors from the different ADNI protocols into a tabular format. 

\subsection{Data Preprocessing}
Variables with missingness at 50\% or greater of the total rows for that predictor were removed. All nominal predictors were dummy-coded. Missing values were imputed using K-Nearest Neighbour imputation with K=5. For the present study, all CSF-derived biomarkers were removed so as to focus on the predictive power of blood-based biomarkers and other variables. Only those who were diagnosed with MCI at baseline were included in the modelling. An additional numeric variable was created indicating the number of months since baseline that the participant had their final visit. For this final visit the participant received a final diagnosis, one of CN, MCI, or AD. These three diagnoses were collapsed into a binary outcome, indicating whether the participant received a worse diagnosis upon final visit, compared to baseline. A worse diagnosis was defined as one that indicated evidence of more severe cognitive decline. Full information on the binary outcomes and their definitions can be seen in Table \ref{tab1}. The final combined dataset contained 2160 variables and 385 observations, including the final visit and final diagnosis variables (Table \ref{tab2}). Full preprocessing and modelling code is available upon request.

\begin{table}[htbp]
	\caption{Those who received a Mild Cognitive Impairment (MCI) diagnosis at baseline were the only
		group included. The models predicted the diagnoses these participants received at the final
		visit, defined here.}
  \centering
	\begin{tabular}{|l|l|l|}
		\hline
  Outcome & Diagnosis \\
  \hline
		CN/MCI      & \makecell{Those diagnosed with MCI at baseline and who received \\ the 
				same diagnosis at their last visit or a diagnosis of CN.} \\
    \hline 
    AD & \makecell{Those diagnosed with MCI at baseline and who received \\ a diagnosis of AD at their
				last visit. }\\
    \hline
	\end{tabular}
	\label{tab1}
\end{table}

\begin{table}[htbp]
	\caption{The final dimensions of the dataset after preprocessing/after ReliefF-based Feature Selection.}
 \centering
 
	\begin{tabular}{|l|l|l|}
		\hline
		Dataset & Variables &  Observations \\
        \hline
		MCI at baseline &  2160/202       & 385 \\ 
		\hline 
	\end{tabular}

	\label{tab2}

\end{table}

\subsection{ReliefF-Based Feature Selection}
Because the dataset contained more variables than observations, which can lead to convergence issues when modelling, it was decided to implement a ReliefF method to select the top 200 most predictive variables in the relation to the outcome. A full description of this method can be found in \cite{urbanowicz_benchmarking_2018} but in brief, the reliefF model cycles through each observation without replacement. For each cycle, the feature score vector $W$ is updated based on feature value differences observed between the target $R_i$ and neighboring instances. ReliefF relies on the user defined parameter $k$ that specifies the use of $k$ nearest hits and $k$ nearest misses in the scoring update for each target instance. It selects $k$ nearest neighbors with the same class called the nearest hits ($H$) and $k$ nearest neighbours of the opposite class, called the nearest misses ($M$). Finally it updates the weight of a feature $A$ in $W$ if the feature value differs between the target instance $R_i$ and any of the nearest hits $H$ or nearest misses $M$ \cite{urbanowicz_benchmarking_2018}.

After undergoing the ReliefF process, the resultant dataset had 200 variables, 385 observations, and the 2 outcome variables indicating final visit and final diagnosis.

\subsection{Models}
\begin{figure}[h]
\centering
\includegraphics[scale=0.145]{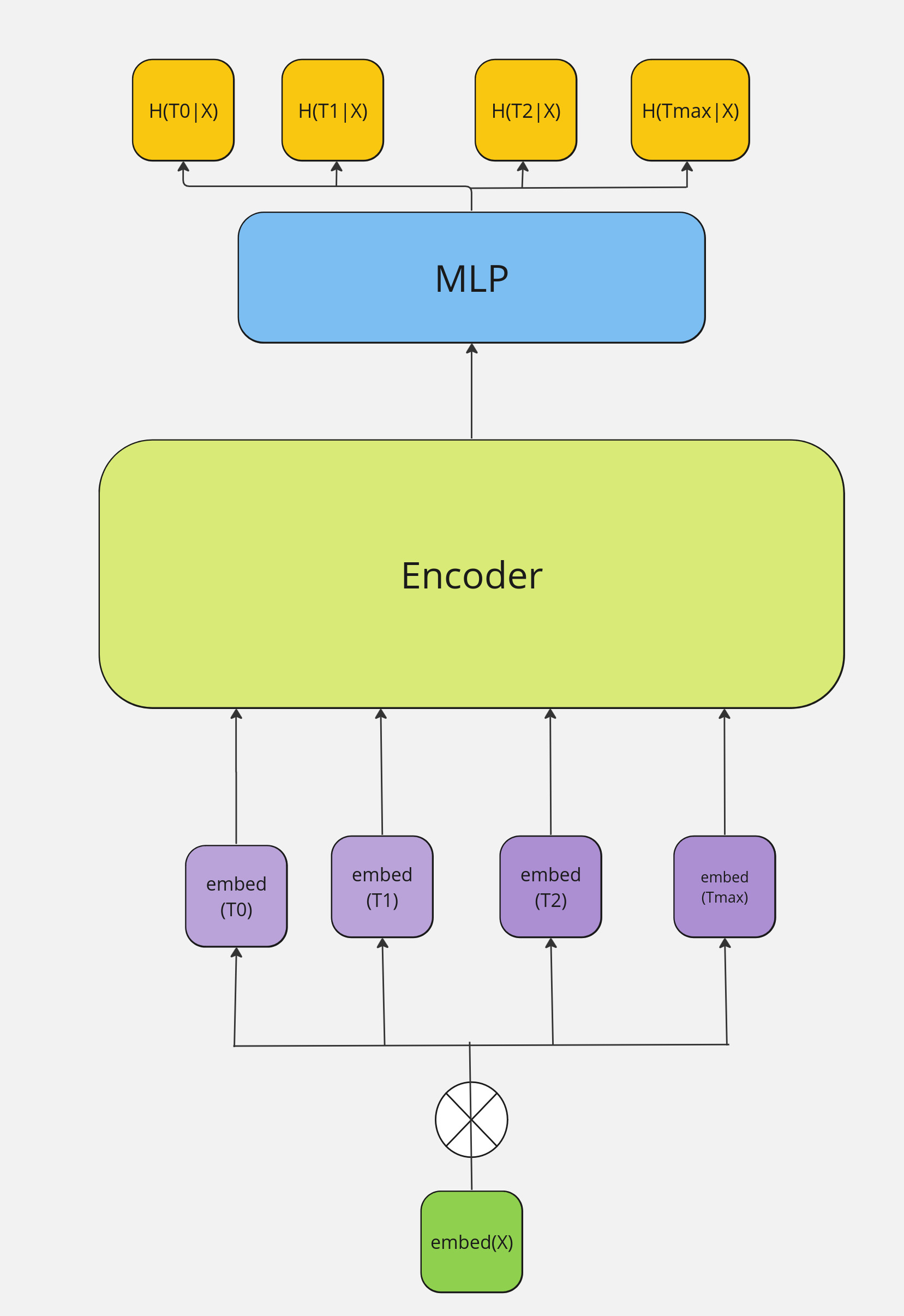}
\caption{A diagram of the architecture of the survival transformer \cite{hu_transformer-based_nodate}.} \label{fig1}
\end{figure}

Model development, evaluation, and validation were carried out according to methodological guidelines outlined by \cite{steyerberg_clinical_2019}; results were reported according to the Transparent Reporting of a multivariable prediction model for Individual Prognosis or Diagnosis (TRIPOD) guidelines \cite{collins_transparent_2015}. This paper explored three survival algorithms:
\\
\\
\begin{enumerate}
    \item Cox Proportional Hazard Model (Cox PH) - The Cox model is expressed by the hazard function, which is the risk of an event occurring at time $t$ as follows:
\\
\noindent
\begin{equation}
h(t) = h_{0}(t) * exp(\beta_{1}X_{1}+\beta_{2}X_{2}+\beta_{p}X_{p})
\end{equation}
\\
where $t$ represents the survival time, $h(t)$ is the hazard function acting upon survival time $t$, $X_{1} , X_{2} , ...X_{p}$ are the values of the $p$ covariates, $\beta_{1} , \beta_{2} , ...\beta_{p}$ are the coefficients that
measure the effect of the covariates on the survival time, and $h_{0}(t)$ is the baseline hazard function. The coefficients are estimated by maximising the partial likelihood and so the model does not require tuning \cite{musto_survival_2023}.
\\
\\
\item Survival XGBoost (SXGB) - Extreme Gradient Boosting is a tree-based ensemble method that grows trees sequentially, by adhering to a gradient descent procedure informed by a loss function, often the negative log liklihood, defined as:
\\
\noindent
\begin{equation}
\min_{\theta} \sum_y {-\log(p(y;\theta))}\label{eq1}
\end{equation}
\\
where the model seeks to minimise the negative log of difference between the true outcome $y$, observed in the training data, and the outcome predicted by the model $\theta$.

This function informs a step function, which calculates the most appropriate adjustments to the model parameters in order to converge on a solution. In the case of the SXGB the negative concordance index is used to calculate steps towards a solution that finds the risk score derived from a Cox Proportional Hazard technique. Thus, the SXGB model seeks to find a risk score that will most closely reflect the true risk for that participant \cite{barnwal_survival_2022}. Hyperparameter tuning was performed on the eta (0.000001 - 0.1), max depth (1 - 5), subsample (0.5 - 0.9), number of columns sampled per tree (0.1 - 0.9), gamma (0.0001 - 0.1), the minimum child weight (1-e8 - 0.0001), and the alpha (0 - 1) and lambda (0 - 20) regularisation parameters.
\\
\\
\item Survival Transformer (STran) - The transformer model, initially proposed for natural language processing tasks, is adapted here for survival analysis for AD, based on work done by \cite{hu_transformer-based_nodate}. At its core, the transformer uses self-attention mechanisms to weigh the significance of different parts of the input data differently, enabling it to capture complex, long-term dependencies. Self attention can be defined as:

\begin{equation}
    \text{Attention}(Q, K, V) = \text{softmax}\left(\frac{QK^T}{\sqrt{d_k}}\right)V
\end{equation}
\\
Where: $Q,K,V$ are the queries, keys, and values respectively, and $d_{k}$ is the dimension of the keys.

The current Survival Transformer employs ordinal regression, treating survival analysis as a problem of predicting ordered categories. This approach allows for directly modeling the survival time distribution as a discrete set of intervals. Ordinal regression is defined as:
\\
\noindent
\begin{equation}
    \log\left(\frac{P(T > t)}{P(T \leq t)}\right) = X\beta - \alpha_t
\end{equation}
\\
Where $t$ is the survival time, $X$ are the input features, $\beta$ are the regression coefficients and $\alpha_{t}$ are the thresholds for the ordinal categories.

An outline of the model architecture can be found in Fig. \ref{fig1}. More information on this model can be found at \cite{hu_transformer-based_nodate}. Hyperparameter tuning was performed on the number of hidden layers (1 - 10) size of the hidden layers in the position-wise feedforward network (1 - 10), the node dropout probability (0.1 - 0.5), the learning rate (1e-6 - 0.01), the regularisation parameters applied to the loss function (0.1 - 3), and the number of epochs (1 - 500). 
\end{enumerate}

\subsection{Bayesian Optimisation}
This paper utlises Bayesian optimisation for the hyperparameter tuning of the SXGB and STran models. Unlike grid or random search methods, Bayesian optimisation uses prior knowledge of the performance of the model with different hyperparameters to intelligently select the next set of hyperparameter values to evaluate. This approach is particularly useful for optimising complex models where training is computationally expensive, as it aims to find the best hyperparameter values in as few evaluations as possible \cite{snoek_practical_2012}.

Bayesian optimisation for hyperparameter tuning operates by intelligently navigating the space of possible hyperparameter settings to find the optimal configuration for a given machine learning model. This process starts with the construction of a surrogate model, typically a Gaussian Process, to estimate the performance across the hyperparameter space. An acquisition function then guides the selection of the next hyperparameters to evaluate by balancing the exploration of new areas against the exploitation of areas known to perform well. The hyperparameters selected by optimising the acquisition function are used to train the model, and the resulting performance metric is fed back into the surrogate model to refine its understanding. This cycle repeats, with each iteration refining the surrogate model's predictions and focusing the search on the most promising regions of the hyperparameter space. This method is particularly valued for its efficiency in finding optimal hyperparameters with relatively few model evaluations, making it suitable for optimizing complex models where training is computationally expensive \cite{shahriari_taking_2016}. For the two models where hyperparameter tuning was available, we tuned the number of rounds of Bayesian optimisation with values 100-1000.

\subsection{Model Performance Indicator}
Performance of the models was assessed using the Concordance index or C-index \cite{steyerberg_clinical_2019}. This metric, also called Harrell’s C-index, provides a global assessment of the model and can be considered a more general form of the AUCROC measure
typically used in binary classification tasks. The C-index computes the percentage of comparable pairs within the dataset whose risk score was correctly identified by the model. More formally:
\\
\noindent
\begin{equation}
C = \frac{\sum_{i,j} I(\hat{t_{i}} > \hat{t_{j}})\cdot I(\eta_{j} > \eta_{i}) \cdot \Delta_{j}} {\sum_{i,j}I(\hat{t_{i}} > \hat{t_{j}})\cdot\Delta_{j}}
\end{equation}

Where $\hat{t_{i}}$ and $\hat{t_{j}}$ are the predicted survival times for individuals $i$ and $j$, respectively, $\eta_{i}$ and $\eta_{j}$ are the actual survival times for individuals $i$ and $j$, respectively, $I\left ( \cdot  \right )$ is an indicator function that returns 1 if the condition inside the parentheses is true and 0 otherwise, and $\Delta_{j}$ is an indicator of whether the event has occurred for individual $j$ (1 if the event has occurred and 0 if censored).
For the full technical details of this index see \cite{schmid_use_2016}.
\\
\subsection{Cross-Validation with Bayesian Optimisation and Monte-Carlo Simulation}
\begin{figure}[h]
\includegraphics[scale=0.35]{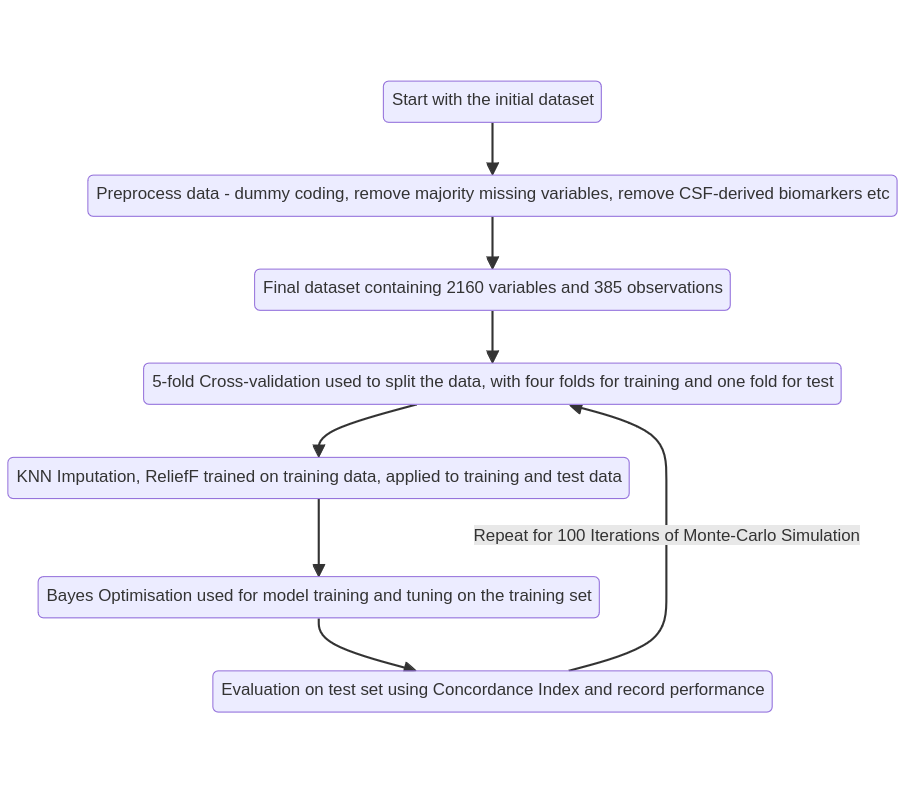}
\caption{Illustration of the complete methodology for preprocessing, training, tuning, and evaluation for this study} \label{fig2}
\end{figure}

A Cross-Validation procedure with Bayesian Optimisation was implemented to tune and evaluate the models so precise estimates of the model’s performance of unseen cases (internal validation) could be gathered \cite{musto_machine_2021}. Cross-Validation consisted of an outer 5-fold CV (model assessment) and an inner Bayesian Optimisation procedure. We conducted a Monte-Carlo procedure of 100 repetitions of the CV+Bayes using different random splits per model to assess the models' stability. Performance statistics were recorded for each model produced by each iteration. Each performance statistic's mean and standard deviation across all iterations were recorded when the Monte-Carlo was complete. To ensure the representativeness of training and test samples in both procedures, the data splitting was stratified based on the AD cases variable. The full methodology can be seen in Fig. \ref{fig2}.
\subsection{Software and Hardware}
The data analysis was conducted using the Python language. Initial data cleaning was performed using the scikit-learn library \cite{noauthor_scikit-learn_nodate}. The reliefF procedure was conducted using the skrebate library \cite{noauthor_skrebate_nodate}. The modelling including training, tuning and evaluation, was performed on the Cox PH model using the lifelines library \cite{davidson-pilon_lifelines_2024}. The SXGB modelling was performed using the xgboost library \cite{noauthor_python_nodate}. For the transfomer modelling, a version of the model found in \cite{hu_transformer-based_nodate} was adapted to be run on ADNI data. The hardware consisted of one server running Ubuntu with a 16-core Ryzen processor, 128 GB of RAM and a 4090 RTX 24GB GPU.

\section{Results}
The C-index performance for each model type performed on the ADNI data is detailed in Table \ref{tab3}.

\begin{table}[htbp]
\caption{C-Index scores for the models applied to the ADNI data on the test set.}
\centering
\begin{tabular}{|l|l|l|}
\hline
Model &  C-index \\
\hline
CoxPH & 0.77 \\
SXGB &  0.80 \\
STran & 0.83 \\
\hline
\end{tabular}
\label{tab3}
\end{table}

The best performing model was the survival transformer, followed by the Survival XGBoost model. Both Survival ML models outperformed the Cox Proportional Hazards model in terms of the C-Index performance metrics.

The Results of the Monte-Carlo Simulation are detailed in Table \ref{tab4}.
\begin{table}[htbp]
\caption{Monte Carlo simulation of 100 iterations for the models applied to the ADNI data on the test set - Mean(SD) of the C-Index.}
\centering
\begin{tabular}{|l|l|l|}
\hline
Model &  Mean C-index(SD) \\
\hline
CoxPH & 0.77 (0.05)\\
SXGB &  0.80 (0.005)\\
STran & 0.85 (0.01) \\
\hline
\end{tabular}
\label{tab4}
\end{table}

\begin{figure}[H]
\centering
\includegraphics[scale=0.65]{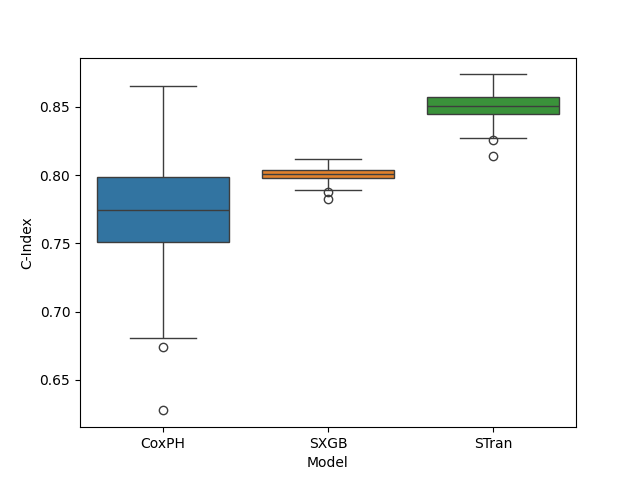}
\caption{Boxplots of C-Index performances for CoxPH, SXGB and STran models obtained
in the Monte Carlo simulation of 100 iterations.} \label{fig3}
\end{figure}

When considering the Monte-Carlo simulation, the survival transformer model proved the best performing in terms of the mean C-index over 100 iterations of the nested cross-validation procedure. See Table \ref{tab4} and and Fig. \ref{fig3} for illustration. In terms of the stability of performance for these models, as measured by the standard deviation of the C-Index over 100 iterations, Survival XGBoost model proved to be the most stable, followed by the survival transformer model. Both Survival ML models were more stable than the standard Cox Proportional Hazards model.

\section{Discussion}
This study aimed to further explore the potential of survival-based ML as a tool for predicting time to AD diagnosis. This paper demonstrates the clear utility of such methods when predicting on the ADNI dataset. This supports the work of \cite{spooner_comparison_2020}\cite{musto_predicting_2023}\cite{musto_survival_2023}\cite{mirabnahrazam_predicting_2023} and thus provides further support for the utility of these survival based techniques in the context of dementia prediction modelling. 

Furthermore, this study also aimed to explore the value of blood-derived biomarkers as predictors in survival-based machine learning and deep learning techniques. Although a number of recent studies have provided evidence for the utility of blood-derived biomarkers in Alzheimer's modelling \cite{teunissen_blood-based_2022}, as far as we are aware, this work is the first to use these biomarkers for survival-based modelling. As has been mentioned by \cite{guo_plasma_2024}\cite{stamate_metabolitebased_2019} these biomarkers would allow a much less invasive way of alerting clinicians and patients to the risk of cognitive decline. If we are also able to predict, not just a binary outcome indicating AD or not, but also the risk of deterioration as a function of time, we are both able to provide more information to clinicians and patients, whilst also providing a better experience through minimally invasive data collection. The current study is therefore a demonstration of good predictive ability, using survival-based techniques and minimally invasive biomarkers. This both lends support for survival based techniques, supporting the work of \cite{musto_machine_2021} and \cite{spooner_comparison_2020}, and for the use of blood derived biomarkers in these models, supporting the work of \cite{stamate_metabolitebased_2019}.

A further novelty for this paper is the use of a survival-based transformer to predict deterioration in the context of AD. Although there is a small but growing body of literature adapting transformers for survival modelling \cite{wang_survtrace_2022}\cite{ding_pathology-and-genomics_2023}\cite{hu_transformer-based_nodate} the present work is the first to apply these techniques to predict deterioration in AD. To that end, this study demonstrated that survival-based transformers have superior performance when compared to survival-based extreme gradient boosting and the standard Cox Proportional Hazards model. This result is notable as previous work which have compared tree-based and neural network-based survival techniques have found that the less complex tree-based methods have performed better than deep learning methods \cite{musto_survival_2023}\cite{musto_predicting_2023}. However, previous work using survival deep learning to predict on deterioration in dementia have focused on recurrent or convolutional neural networks. Survival transformers adopt a wholly different approach, using multiheaded attention and treating each participant as a 'sentence' where each 'word' is the interaction between the participant and time \cite{hu_transformer-based_nodate}. Thus, the transformer attempts to predict the complement of the hazard function for all participants at all points in time \cite{hu_transformer-based_nodate}. Transformers, therefore, represent a new and substantially different approach to all survival based deep learning techniques that have previously been proposed. This may explain why this technique appears superior to the other deep learning techniques presented in the literature. However, transformers are computationally intensive, even when compared to other neural network techniques. If they are to be implemented in health infrastructure and practices, the intensive training required for the models needs to be taken into consideration. Furthermore, inference is difficult with transformers as it is challenging to see how predictions are reached. This means clinicians and researchers are unable to observe potentially modifiable risk factors which increase the risk of AD.

Although this paper presents work with several aspects of novelty, the study contains a number of limitations. Firstly, the Bayesian Optimisation technique was employed for hyperparameter tuning for the survival-based models where appropriate. Due to time and computational limitations, the search for optimal hyperparameters using this technique was not exhaustive and it is possible that subsequent work may improve on the current results using different combinations of hyperparameters.

Secondly, the sample used in this work is comparatively small, with just 385 observations and 200 candidate predictors. It is essential, therefore, that this work be validated on larger datasets before being adopted into healthcare practice and decision making.

Finally, as has been mentioned elsewhere \cite{musto_survival_2023} ADNI is a US based study with roughly 80\% white participants. Future validational work for these models should therefore seek to employ non white western datasets to test how well the models perform on these groups.

\bibliographystyle{splncs04}
\bibliography{ICANN.bib}
\end{document}